\documentclass[letterpaper, 10 pt, conference]{ieeeconf}  

\IEEEoverridecommandlockouts                              

\usepackage{graphics} 
\usepackage{epsfig} 
\usepackage{amssymb}
\usepackage{graphicx}
\usepackage{subcaption}
\usepackage{amsmath}
\usepackage{algorithm}
\usepackage{algorithmic}
\usepackage{bm}

\title{\LARGE \bf
Automatic Mapping with Obstacle Identification for Indoor Human Mobility Assessment
}

\author{V. Ayala-Alfaro, J. A. Vilchis-Mar, F. E. Correa-Tome and J. P. Ramirez-Paredes$^{1}$
\thanks{*This work was supported by Microsoft Corporation as part of the AI4A initiative.}
\thanks{$^{1}$The authors are with the Department of Electronics Engineering,
        University of Guanajuato, Carretera Salamanca - Valle de Santiago km 3.5 + 1.8 Comunidad de Palo Blanco, 36885 Salamanca, Gto. Mexico. Contact: 
        {\tt\small jpi.ramirez@ugto.mx}}%
}

\begin{document}

\maketitle
\thispagestyle{empty}
\pagestyle{empty}

\begin{abstract}

We propose a framework that allows a mobile robot to build a map of an indoor scenario, identifying and highlighting objects that may be considered a hindrance to people with limited mobility. The map is built by combining recent developments in monocular SLAM with information from inertial sensors of the robot platform, resulting in a metric point cloud that can be further processed to obtain a mesh. The images from the monocular camera are simultaneously analyzed with an object recognition neural network, tuned to detect a particular class of targets. This information is then processed and incorporated on the metric map, resulting in a detailed survey of the locations and bounding volumes of the objects of interest. The result can be used to inform policy makers and users with limited mobility of the hazards present in a particular indoor location. Our initial tests were performed using a micro-UAV and will be extended to other robotic platforms.

\end{abstract}

\section{Introduction}

As countries around the globe improve their life expectancy rates and their economies, people are experiencing greater rates of accessibility problems, due in part to the increasing median age of their population and also to occupational hazards. For instance, in Mexico, which is considered an upper-middle income, developing economy \cite{U2007}, the nationwide rate of people with disabilities is 60 per 1000 inhabitants. Out of those people living with disabilities in the country, 64\% have a form of mobility impairment \cite{INEGI2017}.

Many efforts are being conducted to improve the quality of life of people living with disabilities. For instance, goal 11 of the UN Sustainable Development initiative considers the need for public spaces that are inclusive, emphasizing people with disabilities as part of the target community \cite{Nations2015}. In order to reach this goal, countries have to focus on implementing accessibility measures in the design of public sites. 

Despite the need of considering people with mobility impairments during the design of indoor public spaces, a report by the World Health Organization found that, in a survey of 114 countries, more than 40\% did not have any accessibility standards for public buildings \cite{Organization2011}. While this is a changing trend, and more countries are adopting inclusive practices, there is a need for tools that ease the enforcement of accessibility standards. 

An inspector or policy maker can carry out site surveys to assess the accessibility of a given indoor public space. This requires a human expert to devote a significant amount of time on each visit. We propose to automate this task by using a small mobile robot, applying mapping and an object detection algorithms to detect potential hazards or encumbrances to human mobility. 

Many people with mobility impairments use mechanical aids, such as wheeled chairs, walkers, canes, and so on. Designers of indoor public spaces cannot rely on the size data from a single manufacturer of such devices. Instead, there are guidelines and studies that attempt to compile up-to-date statistics on device size \cite{DSouza2010}.

The state-of-the-art for the semantic segmentation of indoor space scenes relies on the use of RGB-D sensors, producing interesting results with a combination of neural networks and random fields, along with other refinements \cite{Tchapmi2017, Pham2019}. These techniques require dense data with direct depth measurements. Instead, we propose to use sparse points from a monocular camera in order to build an indoor map, given that it provides enough precision to describe an indoor scene in terms of its accessibility. In addition, a map that marks the location of obstacles to human mobility should only describe the position and approximate size of these objects, and this information can be computed from 2D images as well. 

The main objective of this work is to develop a solution that, using images from a monocular camera mounted on a mobile robot, extracts relevant points of the scene and constructs a map of the environment, while also detecting and locating objects that can be considered obstacles to human mobility on the same map. The end result is a 3D representation of the environment with enough detail to determine if the indoor space is navigable by users of mobility aids, hinting at which objects could be removed or rearranged to improve accessibility. As we will detail below, we leverage existing simultaneous localization and mapping (SLAM) algorithms, used in conjunction with object detection neural network models, along with our own algorithms for point cloud scaling and the determination of bounding volumes for the obstacles of interest.

\section{Methods} \label{sec:methods}
\subsection{System Description}
\begin{figure}[thpb]
    \includegraphics[width=\columnwidth]{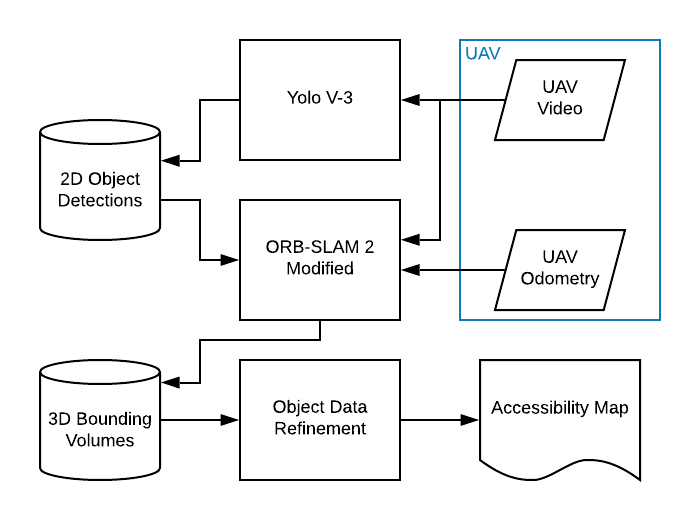}
    \caption{Block diagram of the proposed framework.}
    \label{fig:system_bd}
\end{figure}

The proposed framework employs a mini drone for the initial scanning process of the interest area. Afterward, the video input along with the drone odometry data is used in different stages to detect, label and localize relevant objects in a 2D-3D representation.

The workflow of our framework is divided into 4 main stages, the first related with the objects detection, the next two related with the positioning for both the robot and the objects in a fixed reference frame, and the last one concerning the creation of a user-friendly accessibility map. A visual representation of the main structure of the described system is shown in Figure~\ref{fig:system_bd}.

For the object detection, the YOLO system \cite{yolov3} has been used due to its capability to provide the bounding boxes of the detected objects. Concerning the robot navigation, we used the state-of-the-art ORB-SLAM 2 system \cite{mur2017orb} for its capability to generate a point cloud from a monocular camera and localization for  the drone in real time.

\subsection{Object detection}\label{sec:obj_det}

We used version 3 of the YOLO object detection framework as a basis for our obstacle detection framework. This version of the object detector is able to handle either static images or video streams. In our application, we used the single-frame mode and tuned the detector to find only those object categories that represented furniture or other movable encumbrances to human mobility. The resulting detector focuses on 16 classes, providing the user with the coordinates of the bounding box inside the image for each detected object.

\subsection{Map creation}\label{sec:map_creation}

We adapted the ORB-SLAM 2 system to include scale information, in order to produce a metric map. The original ORB-SLAM 2 implementation can publish or export a point cloud with minimal modification, but, in the case of monocular video input, its scale is not known. Since the ROS driver for both the DJI Tello and the Parrot Bebop 2 drone allows the user to access sensor data, we extracted the estimated linear velocity of the UAV from this data stream.

We use this information to estimate the displacement between two drone positions. First, the drone position when the ORB-SLAM system starts the search of match-points, and later, the drone position when the map is successfully initialized. 

This displacement information allows us to compare it with the displacement calculated internally by ORB-SLAM and scale the map accordingly at its initialization. In the case that map initialization fails, reference drone positions are restored to adequately match the stages of ORB-SLAM map creation, as described previously.

\subsection{Bounding volumes} \label{sec:bou_vol}

\begin{figure}[tpb]
  \includegraphics[width=\columnwidth]{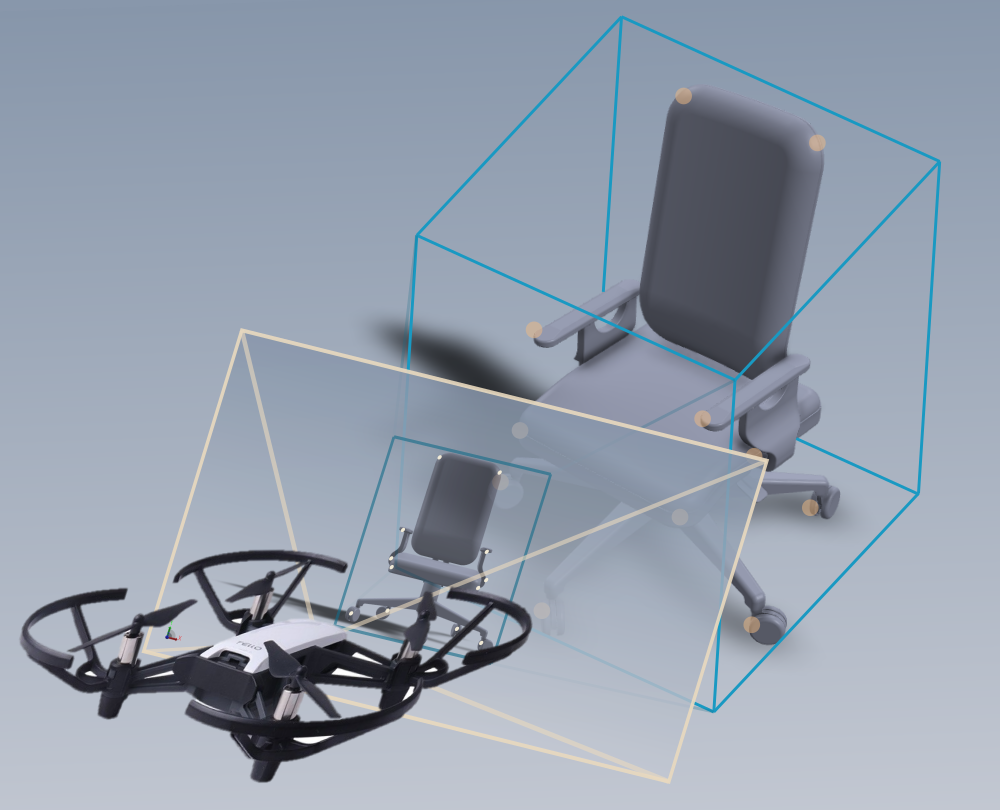}
  \caption{Object detection 2D bounding box converted to 3D bounding volume.}
  \label{fig:2D-3D_conversion}
\end{figure}

We include the information gathered by the object detection module of our system in order to generate bounding volumes. The bounding box data, combined with the point cloud generated by ORB-SLAM is used to project the found bounding boxes from the 2D image plane to a 3D bounding volume as shown in Figure~\ref{fig:2D-3D_conversion}.

From the pinhole camera model, Equation~\ref{eq:pinhole_camera} relates the 3D coordinates of a point, $P$, in the camera field of view, with its matching point in the 2D image plane, $P'$,

\begin{align} 
   & & P'&= KP & & \label{eq:pinhole_camera}\\
   P'&=\begin{bmatrix} u \\ v \\ 1 \end{bmatrix} & 
   K &= \begin{bmatrix}
    f_x     & 0     & c_x   & 0 \\
    0       & f_y   & c_y   & 0 \\
    0       & 0     & 1     & 0
    \end{bmatrix} & P&=\begin{bmatrix} x \\ y \\ z \\ 1 \end{bmatrix} \nonumber
\end{align}

where $K$ is the intrinsic camera matrix, containing the parameters: the focal lengths, $f_x, f_y$, and optical center ,$c_x, c_y$.

\begin{figure}[tpb]
  \centering
  \includegraphics[width=0.8\columnwidth]{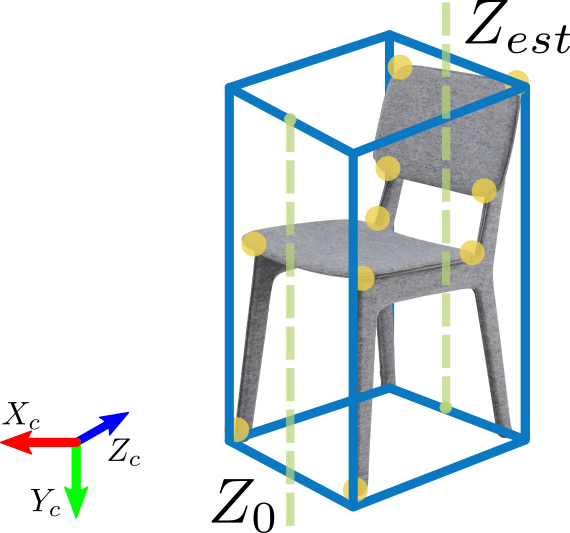}
  \caption{Estimated depth values in camera coordinates.}
  \label{fig:depth_estimation}
\end{figure}

With the aim of projecting the bounding box corners to the 3D space, we solve Equation~\ref{eq:pinhole_camera} for the $x, y$ coordinates. Although Equation~\ref{eq:2D_to_3D} provides a width and height of the object in real world coordinates, it requires a known depth value to project the 2D points to 3D. In order to form a rectangular prism shape we approximate two depth values, corresponding the front ($z = Z_0$) and rear ($z = Z_{est}$) planes of the bounding volume as shown in Figure~\ref{fig:depth_estimation}. 

\begin{equation}\label{eq:2D_to_3D}
  \begin{aligned}
    x = (u - c_x)\frac{z}{f_x}\\
    y = (v - c_y)\frac{z}{f_y}
  \end{aligned}
\end{equation}

To approximate these depth values we make some considerations. The front depth value, $Z_0$, is obtained from the point cloud data by taking the $Z_c$ coordinate of the closest detection inside the detected bounding box. The rear depth value, $Z_{est}$, is approximated as the average of the width and height values, this proves to be an accurate-enough approximation for the possible detected objects considered in this work. 

Nevertheless, the obtained 3D points must be transformed to the global reference frame, $\mathcal{M}$. Hence we use the known transformation of the camera frame, $\mathcal{C}$, w.r.t the UAV frame, $\mathcal{U}$ and the UAV pose information estimated by ORB-SLAM to calculate the global position of the projected bounding box corners, as given by Equation~\ref{eq:camera_to_world},

\begin{align} \label{eq:camera_to_world}
    P_{\mathcal{M}} = {}^{\mathcal{M}} T_{\mathcal{U}} {}^{\mathcal{U}}T_{\mathcal{C}} P
\end{align}

where $P$ is a 3D point in the camera frame, ${}^{\mathcal{U}} T_{\mathcal{C}}$ is the transformation from the camera frame to the UAV frame, ${}^{\mathcal{M}} T_{\mathcal{U}}$ is the transformation from the UAV frame to the global (inertial) frame and $P_{\mathcal{M}}$ is the 3D point expressed w.r.t the global frame.


With all three dimensions (and their vertices positions) estimated, a bounding volume is generated. All points wrapped in this volume are associated to the detected object for further processing. In the case multiple bounding volumes overlap each others, the shared points keep references to all objects they are enclosed in.

\subsection{Accessibility map generation} \label{sec:acc_map}

The 3D spatial object information retrieved (i. e. the bounding volume data) is refined to create an accessibility map. The gathered data includes all the objects appearances through time, together with false detections, which should be purged in order to acquire a user-friendly map.

We propose a refinement of the bounding volumes in 3 main stages: first we check for the reliability of the acquired data discarding obvious faulty detections. Thereupon we try to merge the bounding volumes by comparing detections against each others. Finally, we check the number of frame appearances of the merged objects in order to retain (or remove) them from the final set. 

The proposed procedure is described in Algorithm~\ref{alg:am_generation} with some of the functions further described hereafter:

\begin{itemize}
    \item The \textbf{ValidObject} function prevents having anomalies in the bounding volumes processed. It compares the volume value (size) with minimum and maximum references in order to label the detection as a valid object.
    
    \item The \textbf{VolumeContains} function have an error margin to consider if an detection is contained (or not) in another one. We also check that volume ration between both objects does not exceed a fixed value to avoid merging different instances of one type of objects (e. g. merging two different chairs into one big chair). The function also checks class similarity between the objects to be compared.
    
    \item The \textbf{MergeVolumes} function takes into consideration the number of appearances the objects have associated and adds them up, with the aim of retaining the number of appearances of the object through time. The merged volume keeps the dimensions of the bigger object.
\end{itemize}

\begin{algorithm}[tbp]
\begin{footnotesize}
\begin{algorithmic}[1]
\REQUIRE V \COMMENT{Bounding volumes vector}\\
\ENSURE V\textsuperscript{*} \COMMENT{Refined bounding volumes vector}\\
V\textsubscript{0} = []
\FORALL{v in V}
    \IF{ValidObject(v)}
        \STATE V\textsubscript{0}$.$insert(v)
    \ENDIF
\ENDFOR \\
V\textsubscript{1} = []
\FORALL{v in V\textsubscript{0}}
    \FORALL{v'!= v in V\textsubscript{0}}
    \IF{v$.$VolumeContains(v') \OR v'$.$VolumeContains(v)}
        \STATE V\textsubscript{1}$.$insert(MergeVolumes(v, v'))
        \STATE V\textsubscript{0}$.$remove(v')
    \ENDIF
    \ENDFOR
\ENDFOR \\
V\textsubscript{2} = V\textsubscript{1} 
\FORALL{v in V\textsubscript{2}}
    \IF{v$.$Appearances $<$ app\textsubscript{min}}
        \STATE V\textsubscript{2}$.$remove(v)
    \ENDIF
\ENDFOR
\RETURN V\textsuperscript{*} = V\textsubscript{2}
\end{algorithmic}
\end{footnotesize}
\caption{Bounding volume refinement}
\label{alg:am_generation}
\end{algorithm}

\section{Results}
In order to test the proposed methodology experiments were performed in office-like indoor scenarios. All experiments were executed in a laptop computer running Ubuntu 18.04 and ROS melodic. Regarding the robotic hardware, our tests were conducted using miniature consumer drones: a DJI Tello \cite{tello_driver} and a Parrot Bebop 2 \cite{bebop_driver}. 

Flights were performed in our lab facilities using a manual controller. The interest area was scanned with the aim of including corridors and furniture which may be blocking the mobility of people accessing the lab. Video and odometry data acquired during these flights was stored for subsequent processing, as described in Section~\ref{sec:methods}. 

\subsection{YOLO/ORB-SLAM integration}

\begin{figure}[tpb]
	\centering
	\subfloat[\centering Keypoints detection by ORB-SLAM and object detection from YOLO.]{\includegraphics[width = 0.9\columnwidth]{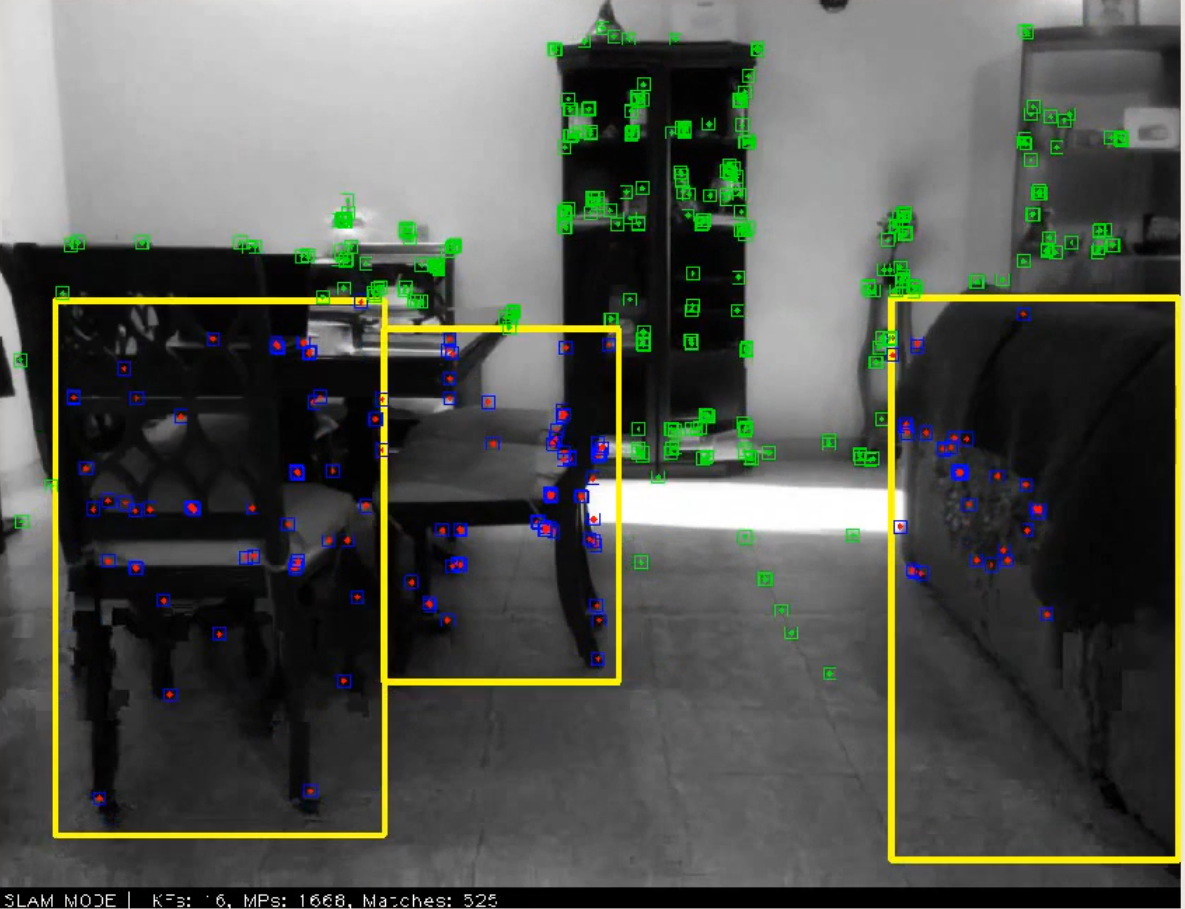}\label{fig:ORB_w_YOLO_ui_a}}
	\\
	\subfloat[Bounding volumes detected in point cloud.]{\includegraphics[width = 0.9\columnwidth]{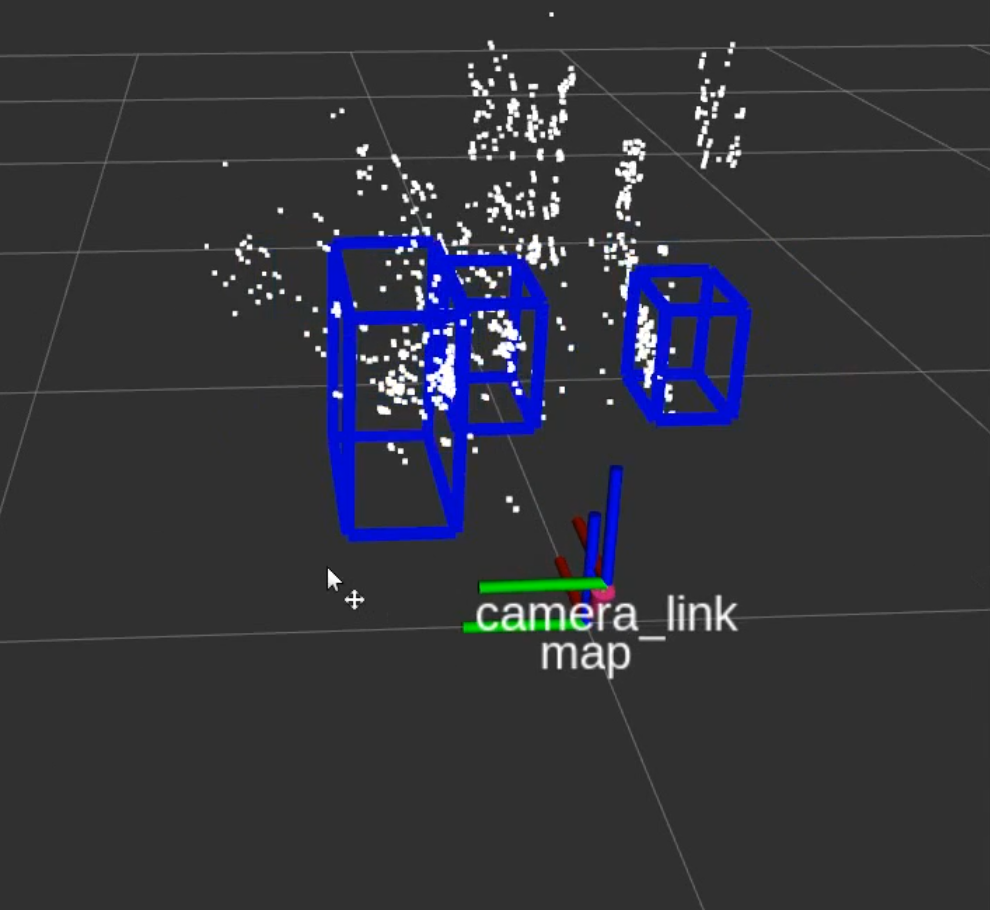}\label{fig:ORB_w_YOLO_ui_b}}
	 \caption{ORB-SLAM modified to include object detection and bounding volumes generation.}
	 \label{fig:ORB_w_YOLO_ui}
\end{figure}

The bounding boxes of the detected objects in the flight video are stored in a CSV file, and afterwards are used during the SLAM procedure. We have modified the ORB-SLAM user interface to include the bounding boxes data generated by the YOLO framework. Besides, the bounding volumes generated (as described in Section~\ref{sec:bou_vol}) are also rendered in real time while generating the point cloud generated by ORB-SLAM. A visualization of these characteristics is shown in Figures \ref{fig:ORB_w_YOLO_ui_a} and \ref{fig:ORB_w_YOLO_ui_b}, respectively.

\subsection{Accessibility map}

Once the detected objects 3D information (i. e. the bounding volumes) is generated, we perform the post-processing as described in Section~\ref{sec:acc_map}. Quantitative and qualitative results obtained concerning the accessibility maps are presented and further discussed in this Section.

\begin{figure}[tpb]
	\centering
	\subfloat[\centering Accessibility map generated with a Tello drone.]{\includegraphics[width = \columnwidth]{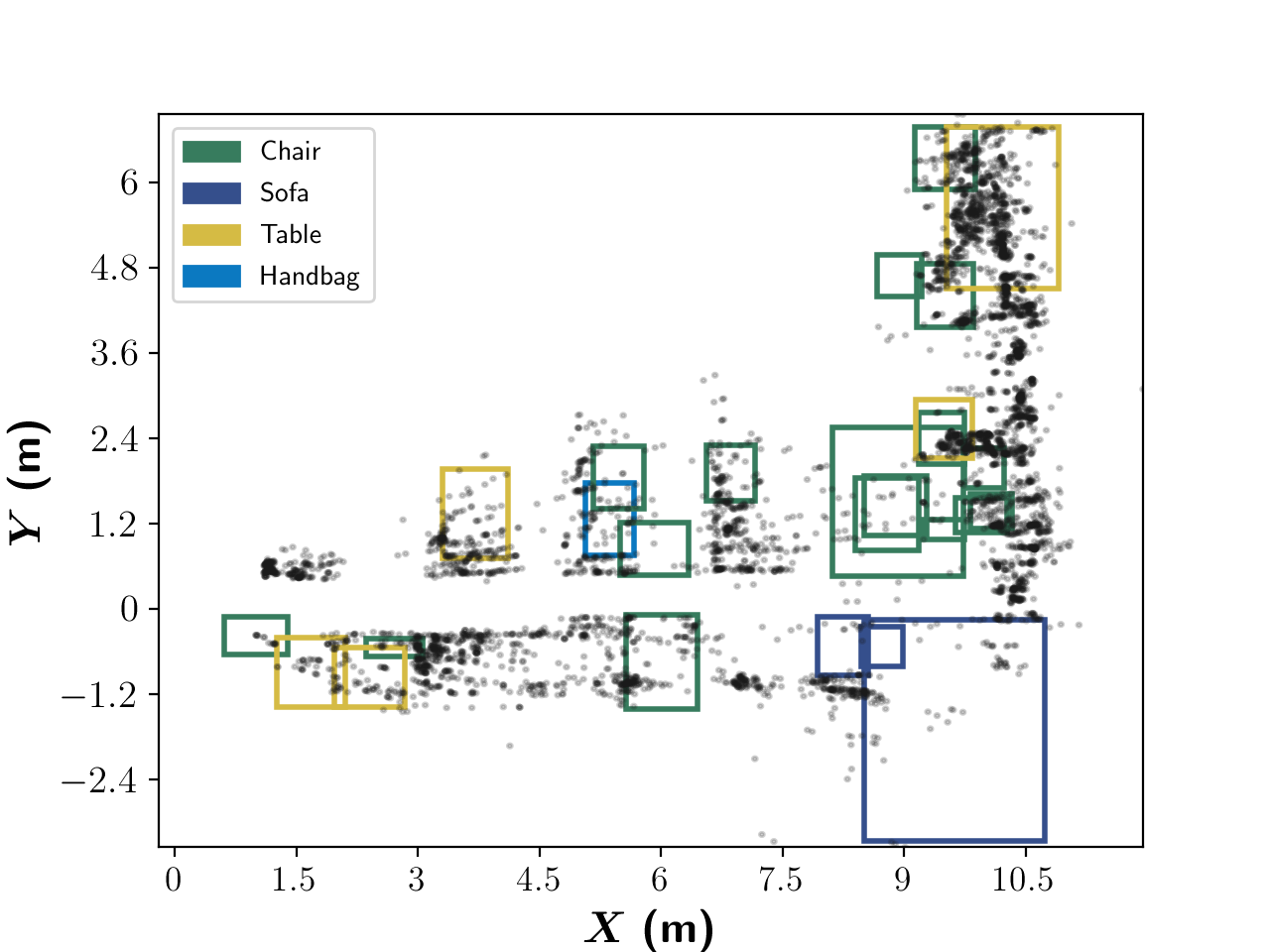}\label{fig:tello_test}}
	\\
	\subfloat[\centering Accessibility map generated with a Bebop 2 drone.]{\includegraphics[width = \columnwidth]{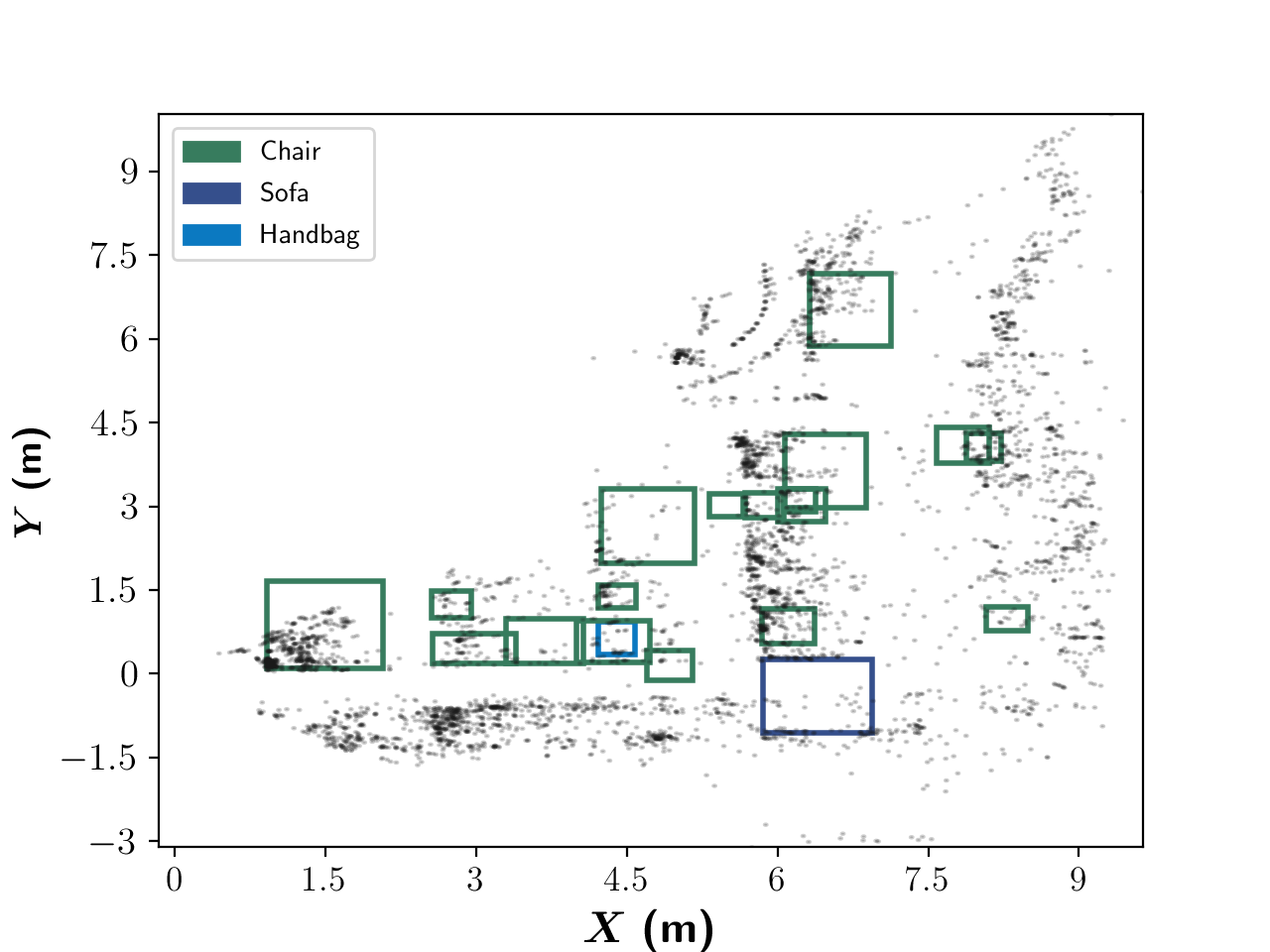}\label{fig:bebop_test}}
	 \caption{Comparison of Accessibility maps generated from two different flights in the same area.}
	 \label{fig:drone_tests}
\end{figure}

In Figure~\ref{fig:drone_tests} a visual comparison between two different scans (performed with different drones) of a common area is presented. In the accessibility map shown in Figure~\ref{fig:tello_test}, multiple object classes are detected. Nonetheless, our system struggles to merge correctly some of the detected objects. Moreover, in the accessibility map shown in Figure~\ref{fig:bebop_test}, less classes are detected correctly, even so, the detected objects are merged more consistently than in the previous case. The difference in the covered area is explained by the fact that flights are performed manually. With the purpose of further improving our system, an autonomous flight plan is in development, to compare maps generated from exact flight replicas.

\begin{figure*}[thbp]
    \centering
    \begin{subfigure}[b]{0.475\textwidth}
        \centering
        \includegraphics[width=\textwidth]{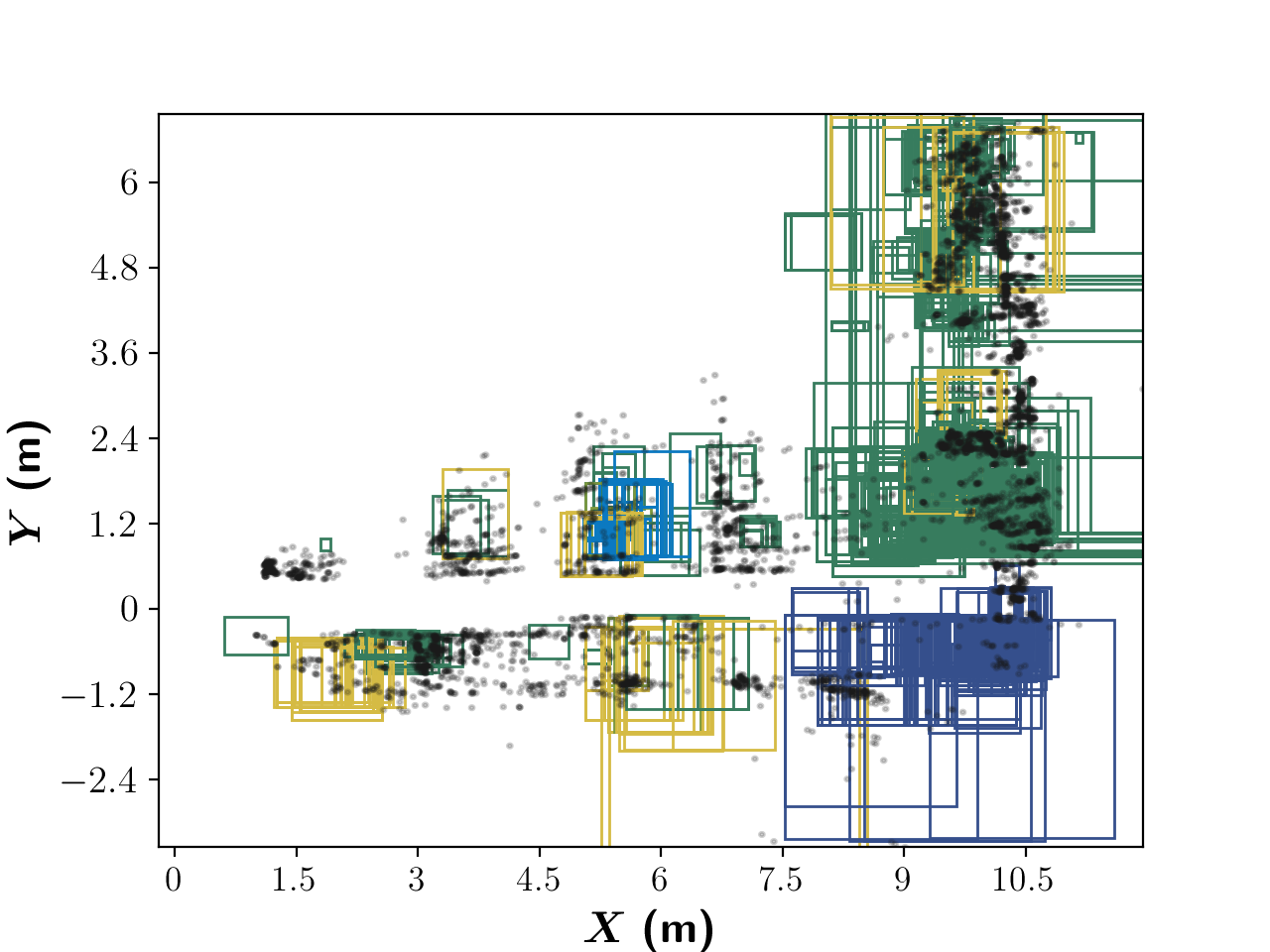}
        \caption[]%
        {Bounding volumes extracted from video.}    
        \label{fig:acc_map_gen_s0}
    \end{subfigure}
    \hfill
    \begin{subfigure}[b]{0.475\textwidth}  
        \centering 
        \includegraphics[width=\textwidth]{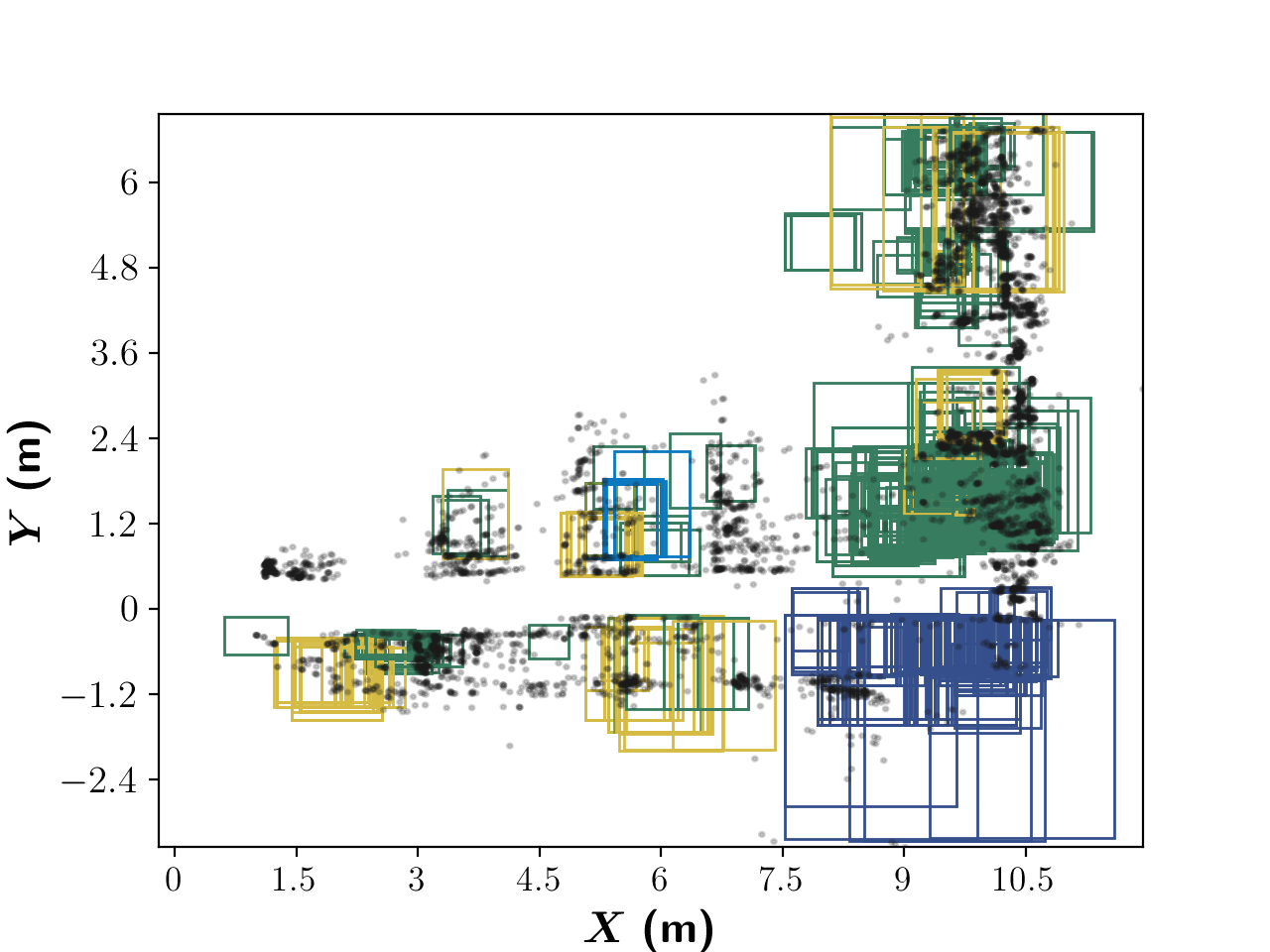}
        \caption[]%
        {Stage 1 (Erroneous detections removing).}    
        \label{fig:acc_map_gen_s1}
    \end{subfigure}
    \begin{subfigure}[b]{0.475\textwidth}   
        \centering 
        \includegraphics[width=\textwidth]{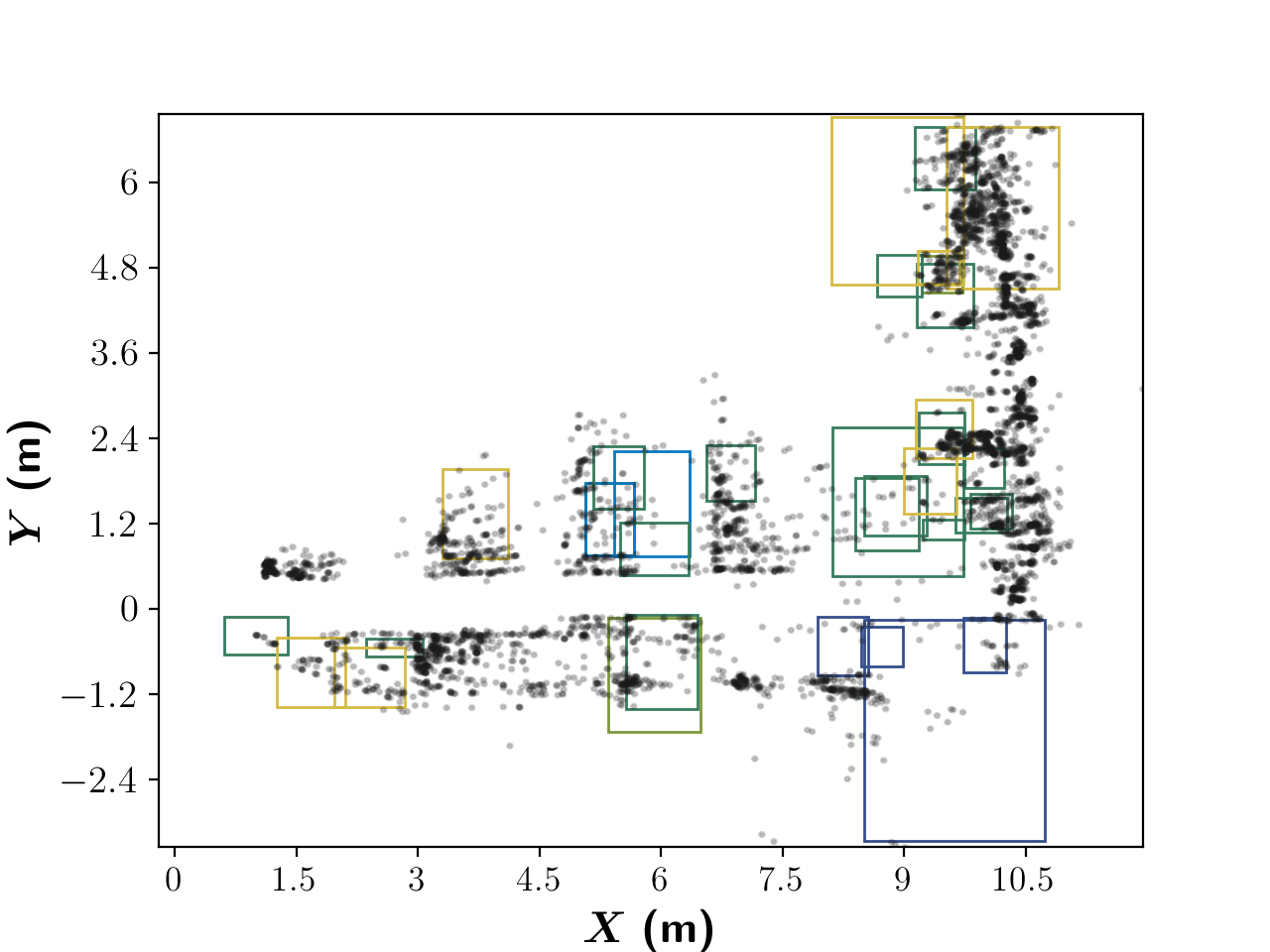}
        \caption[]%
        {Stage 2 (Object merging).}    
        \label{fig:acc_map_gen_s2}
    \end{subfigure}
    \hfill
    \begin{subfigure}[b]{0.475\textwidth}   
        \centering 
        \includegraphics[width=\textwidth]{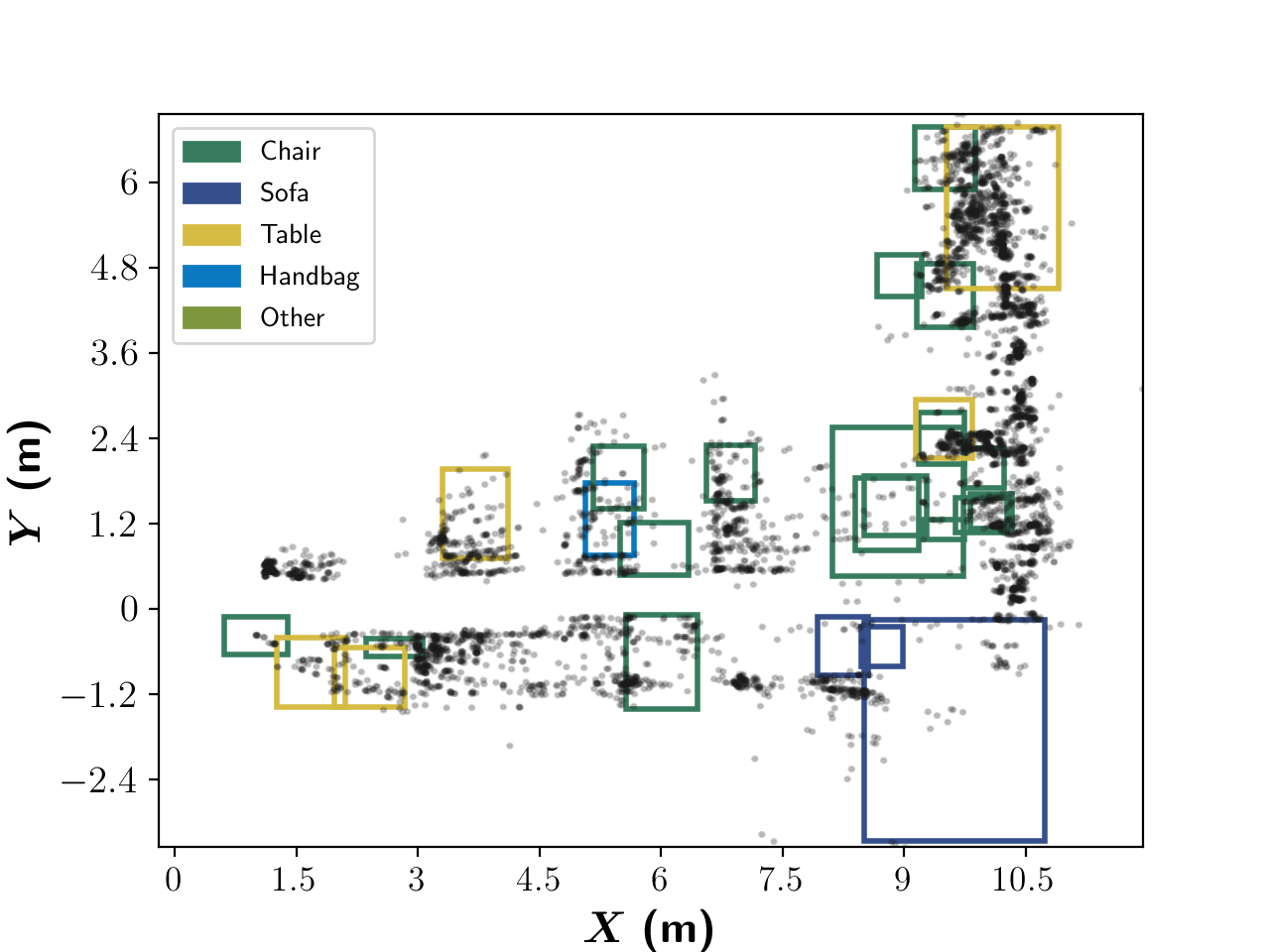}
        \caption[]%
        {Final view of the accessibility map.}    
        \label{fig:acc_map_gen_s3}
    \end{subfigure}
    \caption[]
    {Accessibility map appearance through the main stages of the bounding volumes refinement process.
    } 
    \label{fig:acc_map_generation}
\end{figure*}

Regarding the refinement procedure itself, Figure~\ref{fig:acc_map_generation} shows the process of object refinement for a particular flight is shown. From the raw bounding volumes obtained from the object detection and SLAM algorithm (Figure~\ref{fig:acc_map_gen_s0}) passing through the different Stages of the proposed methodology (Figures~\ref{fig:acc_map_gen_s1}-\ref{fig:acc_map_gen_s3}). Results presented in Table~\ref{tab:acc_map_results} show a significant decrease in the number of objects due to the refinement process. The first Stage should not decrease the number of object appearances considerably given a stable video input, however, this proves to be wrong in the tests performed with the Tello drone where a significant portion of the bounding volumes detected is discarded. As intended, the main contribution to the reduction in the number of objects occurs in the merge phase (Stage 2), due to the shape and size similarity from same objects appearing multiple times through the video frames. In comparison, Stage 3 provides a minor decrease in object appearances, however it accomplishes its function as a filter for wrong classifications by the  object detector.

Some difficulties were encountered by using the proposed hardware. Particularly, the Tello driver produces some noisy images while using the onboard camera, increasing the challenge for the accessibility map generation. Moreover, even if the Bebop 2 driver generates more stable video inputs, a smaller image size is available for use, diminishing the quality of the object detector output. In conclusion, due to the stability provided in the video feed, the Bebop 2 drone is better suited for the map generation task.

\begin{table}[thbp]
\caption{Objects detected in accessibility maps generated} %
	\begin{center}
		\begin{tabular}{|c|c|c|c|c|}
    		\hline
			\textbf{}&\multicolumn{4}{|c|}{\textbf{Objects detected}} \\
    		\cline{2-5}
    		\textbf{Test ID} & \textbf{Stage 0} & \textbf{Stage 1} & \textbf{Stage 2} & \textbf{Stage 3}\\
    			\hline
			Tello test 1 & $1937$ & $902$ & $34$ & $26$\\
			\hline
			Tello test 2 & $524$ & $362$ & $26$ & $20$\\
			\hline
			Bebop test 1 & $1354$ & $1168$ & $28$ & $16$\\
			\hline
			Bebop test 2 & $1082$ & $885$ & $21$ & $20$\\
			\hline
		\end{tabular}
		\label{tab:acc_map_results}
	\end{center}
\end{table}


\section{Conclusions}
A methodology to build an accessibility map of an indoor scenario by scanning the area using a UAV has been proposed and evaluated. Different tests had been performed to prove the viability of this approach with promising results. Based on the obtained results, the accuracy of the detected objects is limited by the quality of the object detector, hence, the use of a more precise detection framework would be necessary to improve the results of the proposed system. 

As future work, there are several ways to increment or improve the presented work. Some work could be done in order to generate a 3D accessibility map (instead of 2D maps), however it may prove to be more complicated to interpret by the intended user. Besides, an automated exploration routine for the UAV is in development for better comparisons between different data acquisition robots. Regarding the accessibility map refinement, more stages could be added with the purpose of improving the merging between different object appearances.

\section*{Acknowledgements}

V. Ayala-Alfaro thanks the Mexican Council for Science and Technology (Conacyt) for scholarship \#1042278. J. A. Vilchis-Mar thanks the University of Guanajuato for a student research grant during the first semester of 2021.

\bibliography{amaiusrefs}
\bibliographystyle{IEEEtran}

\end{document}